\documentclass[12pt]{article}

\usepackage{amssymb}
\usepackage{amsmath}
\usepackage{txfonts}
\usepackage{multirow}
\usepackage{threeparttable}
\usepackage{caption}
\usepackage{tabularray}
\usepackage{authblk}
\usepackage{graphicx}
\usepackage{listings}
\usepackage[top=25mm, bottom=30mm, left=25mm, right=25mm]{geometry}

\usepackage{hyperref}

\usepackage[
citestyle = numeric-comp,
bibstyle=trad-abbrv,
sorting = none,
date = year,
]{biblatex}
\renewcommand{\appendix}{
\setcounter{figure}{0}%
\renewcommand{\thefigure}{S\arabic{figure}}%
\setcounter{section}{0}%
\renewcommand{\thesection}{Appendix \Alph{section}}%
\setcounter{equation}{0}%
\renewcommand{\theequation}{\Alph{section}.\arabic{equation}}
}

\begin{document}
\date{}
\title{Effects of model architecture and learning strategies on deep learning-based recognition of activated sludge microscopic images and comparison with quantitative image analysis}

\author[1]{Suguru Hakoshima}
\author[1,2,3]{Tomohiro Tobino\thanks{t\_tobino@esc.u-tokyo.ac.jp}}
\author[2]{Fumiyuki Nakajima}

\affil[1]{\small\itshape Department of Urban Engineering, Graduate School of Engineering, The University of Tokyo, 7-3-1, Hongo, Bunkyo-ku, 113-8656, Tokyo, Japan}
\affil[2]{\small\itshape Environmental Science Center, The University of Tokyo, 7-3-1, Hongo, Bunkyo-ku, 113-0033, Tokyo, Japan}
\affil[3]{\small\itshape Collaborative Research Institute for Innovative Microbiology, The University of Tokyo, 1-1-1, Yayoi, Bunkyo-ku, 113-8657, Tokyo, Japan}

\maketitle
\begin{abstract}
Microscopic image analysis has long been recognized as a promising approach for monitoring activated sludge. In recent years, deep learning-based image analysis has been increasingly adopted in this field because of its high performance. However, previous studies on microscopic image analysis of activated sludge have rarely explored transformer-based models or self-supervised foundation models and have instead relied on CNNs and supervised ImageNet pretraining. In addition, previous studies often downsampled image sizes, but the effects of downsampling have not been sufficiently investigated, and the relationship between downsampling strategies and image analysis performance remains unclear. Furthermore, no study has quantitatively compared deep learning performance with quantitative image analysis (QIA), which was widely used before the emergence of deep learning. In this study, to examine how model architecture and learning strategies affect performance in microscopic image analysis of activated sludge and to quantitatively determine whether deep learning outperforms QIA, we prepared three types of activated sludge samples, classified their microscopic images, and evaluated classification accuracy. Our results showed that transformer-based architectures and alternative pretraining methods were effective in terms of classification accuracy. Our downsampling analysis showed that using overly small images reduced accuracy, but increasing image size beyond a certain point did not improve it further. In addition, the analysis indicated that, to achieve high classification accuracy, maintaining the field of view was a more effective downsampling strategy than maintaining resolution. Finally, our comparison between deep learning and QIA showed that deep learning outperformed QIA in terms of accuracy.
\end{abstract}

\section{Introduction}

Activated sludge is a biological wastewater treatment process widely used in municipal wastewater treatment plants (WWTPs). It plays a crucial role in preventing water pollution, protecting aquatic ecosystems, and supporting sustainable environmental management by removing organic matter and nutrients from wastewater.
In recent years, image analysis-based artificial intelligence (AI) has achieved remarkable success \cite{Krizhevsky-2012-ImageNetClassificationDeepConvolutional,He-2016-DeepResidualLearningImage,Dosovitskiy-2021-ImageWorth16x16Words,Liu-2022-ConvNet2020s,Siméoni-2025-DINOv3}. In activated sludge systems, AI-driven image analysis has increasingly been explored for process condition monitoring and for the detection and prediction of operational problems. In particular, several studies \cite{Satoh-2021-DeepLearningbasedMorphologyClassification,Borzooei-2024-EvaluationActivatedSludgeSettling,Kaushalya-2025-QuantificationMorphologicalCharacteristicsFilamentous,Bähr-2025-IntroducingSituActivatedSludge,Li-2025-RealtimeQuantificationActivatedSludge,Hakoshima-2024-DevelopmentComprehensiveDetectionAutoclassification} have applied deep learning techniques to microscopic images of activated sludge in order to improve process understanding and control.

Before the widespread application of deep learning to microscopic image analysis of activated sludge, quantitative image analysis (QIA) was the predominant analytical framework. \cite{Mesquita-2013-ActivatedSludgeCharacterizationMicroscopy}.
In typical QIA-based approaches, microscopic images were first binarized, and floc structures were characterized using morphological parameters \cite{Grijspeerdt-1997-ImageAnalysisEstimateSettleability}.
These morphological parameters were subsequently analyzed to investigate their relationships with various operational and performance indicators, including sludge settling performance \cite{Grijspeerdt-1997-ImageAnalysisEstimateSettleability,Amaral-2005-ActivatedSludgeMonitoringWastewater,Mesquita-2009-CorrelationSludgeSettlingAbility,Nakaya-2024-TracingMorphologicalCharacteristicsActivated}, effluent water quality \cite{Mesquita-2016-EstimationEffluentQualityParameters} and dewaterability \cite{Nakaya-2024-TracingMorphologicalCharacteristicsActivated}. Although QIA technologies provided powerful analytical tools, the binarization step was laborious and depended on semi-automated, operator-dependent procedures \cite{Satoh-2021-DeepLearningbasedMorphologyClassification}. Consequently, convolutional neural networks (CNNs), a type of deep learning method, began to be applied to the microscopic image analysis of activated sludge to achieve automated image processing \cite{Satoh-2021-DeepLearningbasedMorphologyClassification}. To date, CNN-based approaches in microscopic image analysis of activated sludge have been reported to enable the estimation of settling performance \cite{Borzooei-2024-EvaluationActivatedSludgeSettling, Kaushalya-2025-QuantificationMorphologicalCharacteristicsFilamentous} and MLSS \cite{Li-2025-RealtimeQuantificationActivatedSludge}, as well as the detection of protozoa \cite{Hakoshima-2024-DevelopmentComprehensiveDetectionAutoclassification,Lei-2024-IdentificationActivatedSludgeMicrobial,Liang-2025-AutomaticVisualDetectionActivated}.


However, previous studies that apply deep learning to the analysis of microscopic images of activated sludge still face several limitations.
First, the architectural design and training strategies of deep learning models remain insufficiently investigated.
Although several studies \cite{Satoh-2021-DeepLearningbasedMorphologyClassification,Borzooei-2024-EvaluationActivatedSludgeSettling} have compared different CNN architectures, transformer-based models \cite{Dosovitskiy-2021-ImageWorth16x16Words} have not yet been systematically evaluated, despite their increasing prominence in recent years.
Moreover, most prior work \cite{Satoh-2021-DeepLearningbasedMorphologyClassification,Borzooei-2024-EvaluationActivatedSludgeSettling,Li-2025-RealtimeQuantificationActivatedSludge} has relied on supervised pre-training on ImageNet, whereas more recent strategies, such as self-supervised learning \cite{Siméoni-2025-DINOv3}, have yet to be thoroughly applied.
In addition, images are generally downsampled before model training because processing large images requires substantial memory \cite{Al-Ani-2026-CrossresolutionLearningScalableDetection}. However, few studies have investigated how image downsampling affects analytical performance, and the choice of resizing strategy has received little attention. Most studies resize images while maintaining the field of view, and no study has systematically compared this approach with downsampling that maintains the resolution.
Comparisons between QIA and deep learning also remain limited. Previous study \cite{Satoh-2021-DeepLearningbasedMorphologyClassification} has described QIA as a semi-automated approach with several limitations and have utilized CNNs as an alternative. Although we agree with this assessment, there has been little experimental evidence that deep learning outperforms QIA in the analysis of microscopic images of activated sludge, particularly in terms of analytical performance.


The objectives of this research were to examine, from the standpoint of analytical capabilities, how various training conditions, including model architecture, pre-training strategies, and microscopic image characteristics, influence deep learning performance, and to compare deep learning and QIA from the same perspective.
We prepared three types of activated sludge samples and constructed a benchmark classification task in which microscopic images obtained from the samples were classified by sample type using deep learning and QIA.
Here, we show that image size affects model performance, whereas model size has little effect. We also demonstrate that models pretrained using self-supervised learning and Vision Transformer (ViT) achieve sufficient performance, that maintaining the field of view in microscopic images is more important than maintaining spatial resolution, and that deep learning outperforms QIA in terms of both analytical performance and the number of images required.


\section{Materials and methods}
\subsection{Sample collection}
Samples were collected from three process lines across two WWTPs in Japan.
The collected samples were stored under refrigerated conditions until microscopic image acquisition.
The detail of samples is shown in \autoref{table:sample}.

\begin{table}[htpb]
\centering
\begin{threeparttable}
\caption{Sample overview}
\label{table:sample}
\begin{tabular}{ccccc}\hline
Sample name & WWTP\tnote{\textit{a}} & Treatment methods\tnote{\textit{b}} & Sampling date\tnote{\textit{c}} &  MLSS (mg/L)\\\hline
Sample 1 & A & AO & 2024-Oct-9th & 1,470 \\
Sample 2 & B & p-AO & 2024-Nov-21st & 1,300 \\
Sample 3 & B & Step A2O & 2024-Nov-21st & 1,510 \\\hline
\end{tabular}
\begin{tablenotes}
\item[\textit{a}] All WWTPs operate under a combined sewer system
\item[\textit{b}] \textit{AO}, Anaerobic-aerobic process; \textit{p-AO}, pseudo anaerobic-aerobic process; \textit{Step A2O}, step-feed anaerobic-anoxic-aerobic process
\item[\textit{c}] The images were acquired on November 28, 2024.
    \end{tablenotes}
\end{threeparttable}
\end{table}


\subsection{Benchmark classification task}
In this study, we set up a benchmark classification task to systematically investigate how deep learning training conditions affect classification performance and to compare deep learning with QIA. In this task, microscopic images were classified by sample, and performance was assessed using classification accuracy (Eq. (\ref{definition-accuracy})) where TP, FP, TN, and FN represent true positives, false positives, true negatives, and false negatives, respectively. We first evaluated the effects of different model size, model architectures, pretraining methods, image sizes, and resizing strategies on the accuracy of deep learning. We then compared the accuracy achieved by deep learning with that achieved by QIA.

\begin{align}\label{definition-accuracy}
\mathrm{Accuracy} = \frac{\mathrm{TP} + \mathrm{TN}}{\mathrm{TP} + \mathrm{TN} + \mathrm{FP} + \mathrm{FN}}
\end{align}



\subsection{Deep learning}
\subsubsection{Image acquisition}
\begin{figure}[hptb]
\centering
\includegraphics[width = 12cm]{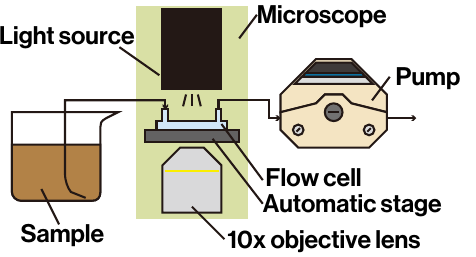}
\caption{Structre of a automated imaging system}
\label{fig:structure}
\end{figure}

A simple automated imaging system (\autoref{fig:structure}) was constructed using a flow cell (Type 48, Starna), an inverted microscope (IX73, Evident), an automatic stage (B30102, CENTRAL MOTOR WHEEL CO., LTD), and a camera (NOA2000, Wraymer, 1824 px $\times$ 1216 px). A 10$\times$ objective lens was used. The imaging procedure was as follows: the sample was placed in a beaker, pumped into the flow cell mounted on the automatic stage using a peristaltic pump, allowed to settle for approximately 1 min, and imaged across multiple fields of view (25 fields of view, generally)  by driving the automatic stage.
In total, more than 300 images were acquired per sample.

\subsubsection{Software and workstation}
Python 3.11 was used as the programming language for the implementation of deep learning.
The following packages were employed: PyTorch 2.7.1, torchvision 0.22.1, timm 1.0.19, and grad-cam 1.5.5.
In addition, the official implementation of DINOv3 \cite{Siméoni-2025-DINOv3} provided by its authors was used.
The prepared source code was executed on a workstation equipped with an AMD Ryzen Threadripper 7970X CPU (Advanced Micro Devices, Inc.) and an NVIDIA RTX 5090 GPU (NVIDIA Corporation).
The virtual environment was constructed using Docker.

The obtained results were analyzed using R 4.4.2. The effsize package (version 0.8.1) was used to calculate effect sizes.

\subsubsection{Models}
In this study, we did not train models from scratch; instead, we retrained eight pretrained models (\autoref{table-modelzoo}).
To compare CNN and Transformer architectures, we employed ConvNeXt \cite{Liu-2022-ConvNet2020s} and ViT \cite{Dosovitskiy-2021-ImageWorth16x16Words,Darcet-2023-VisionTransformersNeedRegisters}, respectively. ViT is a widely used Transformer architecture, whereas ConvNeXt is a modern CNN architecture designed using principles inspired by Vision Transformers, making it suitable for comparison with ViT \cite{Liu-2022-ConvNet2020s}.
We used the ViT implementation developed in a previous study \cite{Siméoni-2025-DINOv3}. This implementation supports adjustable input image sizes and incorporates register tokens \cite{Darcet-2023-VisionTransformersNeedRegisters}, which were not included in the original ViT architecture \cite{Dosovitskiy-2021-ImageWorth16x16Words}.
We also compared different pretraining methods. Most previous studies on microscopic image analysis of activated sludge \cite{Satoh-2021-DeepLearningbasedMorphologyClassification,Borzooei-2024-EvaluationActivatedSludgeSettling,Li-2025-RealtimeQuantificationActivatedSludge,Wang-2024-PredictionActivatedSludgeSedimentation} have used models pretrained through supervised learning on the ImageNet-1K \cite{Deng-2009-ImageNetLargescaleHierarchicalImage} image classification dataset. In recent years, however, self-supervised learning has become an increasingly common approach to model pretraining \cite{Siméoni-2025-DINOv3,He-2020-MomentumContrastUnsupervisedVisual,Chen-2020-SimpleFrameworkContrastiveLearning,Chen-2020-ExploringSimpleSiameseRepresentation}.
Therefore, we compared ConvNeXt models pretrained through supervised learning on ImageNet-1K with ConvNeXt models distilled \cite{Hinton-2015-DistillingKnowledgeNeuralNetworka} from the foundation model pretrained using self-supervised learning in the DINOv3 project \cite{Siméoni-2025-DINOv3}.
Furthermore, to investigate the effect of model size, we employed multiple ConvNeXt and ViT variants with varying numbers of parameters.
\begin{table}[tb]
\centering
\caption{Eight deep learning models used in this study}
\label{table-modelzoo}
\centering
\begin{tblr}{
    hline{1,2,Z} = {0.08em},
    colspec = { X[9,c]X[4,c]X[4,c]X[3,c]},
    cell{2}{2} = { r = 6, c = 1 }{ halign = c, valign = m },
    cell{2}{3} = { r = 3, c = 1 }{ halign = c, valign = m },
    cell{8}{2} = { r = 2, c = 1 }{ halign = c, valign = m },
    cell{5}{3} = { r = 5, c = 1 }{ halign = c, valign = m },
}
Model name & Architecture & Pretraining method  &\# Params (M)\\
ConvNeXt-Tiny\_ImageNet  & {ConvNeXt \cite{Liu-2022-ConvNet2020s}\\(CNN)} & {Supervised learning\\(ImageNet1k)} & 29\\
ConvNeXt-Small\_ImageNet  & ConvNeXt & & 50\\
ConvNeXt-Base\_ImageNet  & ConvNeXt & & 89\\
ConvNeXt-Tiny\_DINOv3  & ConvNeXt & {Self-supervised learning\\(DINOv3\cite{Siméoni-2025-DINOv3})}& 29\\
ConvNeXt-Small\_DINOv3  & ConvNeXt & &  50\\
ConvNeXt-Base\_DINOv3  & ConvNeXt & & 89\\
ViT-Small\_DINOv3 & {ViT \cite{Dosovitskiy-2021-ImageWorth16x16Words,Darcet-2023-VisionTransformersNeedRegisters}\\(Transformer)} & & 21\\
ViT-Base\_DINOv3 & ViT & & 86\\
\end{tblr}
\end{table}


\subsubsection{Training}
The eight models were trained to classify our microscopic images by samples. Training was conducted under multiple conditions, resulting in a total of 48 trained models.
For each sample, 201 -- 207 images were used for training and 100 -- 107 images were used for evaluation. Cross-entropy loss and the Adam optimizer were used in all 48 training runs. The learning-rate schedules are shown in Fig. S1.

To investigate the effects of training conditions on model performance, the models were trained under multiple conditions. Data augmentation was performed as follows. First, vertical and horizontal flips were independently applied with probabilities of 50\%, followed by brightness adjustment. In accordance with a previous study \cite{Borzooei-2024-EvaluationActivatedSludgeSettling}, brightness was varied within a range of 20\%. The augmented images were then resized using PyTorch's \texttt{RandomResizedCrop} function. To evaluate the effect of image size, three input sizes were examined: 256 $\times$ 256, 672 $\times$ 672, and 896 $\times$ 896 pixels. Following the normalization approach used by previous research \cite{Borzooei-2024-EvaluationActivatedSludgeSettling}, the images were normalized using channel-wise means of 0.485, 0.456, and 0.406 and standard deviations of 0.229, 0.224, and 0.225 for the red, green, and blue channels, respectively.
For each image size, two training strategies were applied to investigate the effect of layer unfreezing. In the first strategy, only the fully connected layer (FC layer \cite{Gao-2025-AIDrivenEarlyWarningSludge}) was unfrozen, whereas in the second strategy, all layers were unfrozen. Batch sizes of 8 and 4 were used for the former and latter strategies, respectively. Applying the two training strategies to each combination of eight models and three input image sizes resulted in a total of 48 trained models.

\subsubsection{Evaluation}
We evaluated the 48 models using two resizing strategies to determine whether maintaining the field of view or maintaining resolution resulted in higher classifcation accuracy.
Under the former strategy, the central 1216 $\times$ 1216 pixel region of each test image was cropped and then resized to the input image size used for model training. The resized images were subsequently normalized using the same procedure as that applied during training. Under the latter strategy, the central region of each test image was cropped directly to the input image size used for model training. The cropped images were subsequently normalized using the same procedure as that applied during training.
When examining the effects of factors other than the resizing strategy, we used the evaluation results obtained with the former strategy because maintaining the field of view is more commonly used and has also been adopted in previous study \cite{Al-Ani-2026-CrossresolutionLearningScalableDetection}.

\subsection{Quantitative image analysis}

\subsubsection{Image acquisition}
Microscopic images for QIA were acquired using a different method from that used for deep learning because all flocs needed to be in focus for accurate calculation of morphological parameters.
A 50 \textmu L aliquot of the sample was placed onto a glass slide and covered with a 24 mm $\times$ 24 mm coverslip. The slide was examined using a microscope (SZX16, Olympus) at 10x magnification, equipped with a camera (NOA2000, Wraymer). From each slide, 50 images were captured.

\subsubsection{Software}
All image processing and morphological parameter calculations were performed in MATLAB 2024b.
All other analyses were conducted in R 4.4.2,
and machine learning models were implemented using the  tidymodels 1.2.0, bonsai 0.4.0, and DALEX 2.4.3 packages.

\subsubsection{Image binarization and calculation of morphological parameters}
\begin{figure}[hbt]
\centering
\includegraphics[width = 8cm]{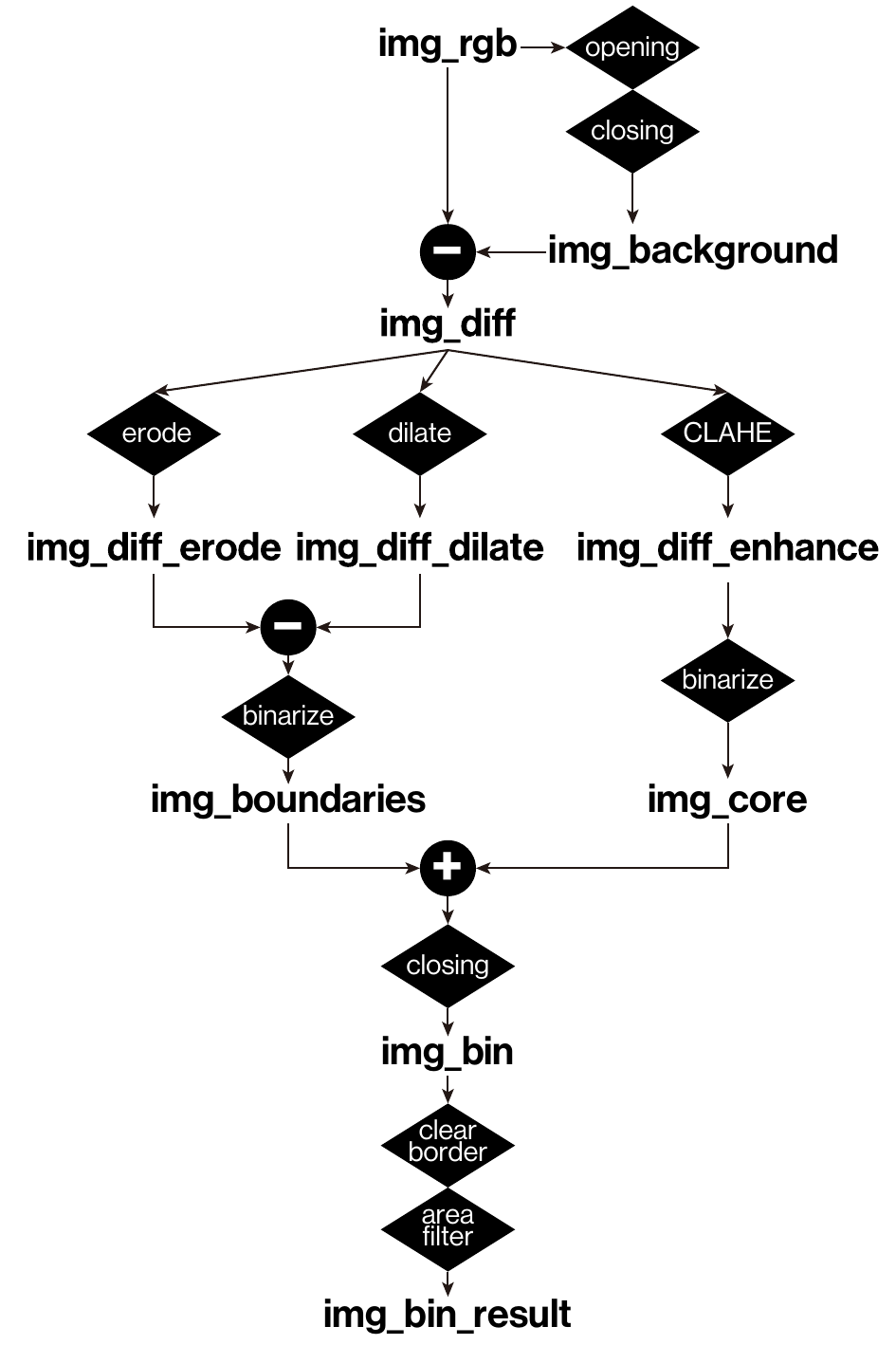}
\caption{The binarization procedure for quantitative image analysis}
\label{figure-binarization-concept}
\end{figure}

The binarization procedure is shown in \autoref{figure-binarization-concept}.
First, a background image was generated from the initial microscopic image using morphological opening and closing (\autoref{figure-binarization-concept}, img\_background) \cite{daMotta-2001-CharacterisationActivatedSludgeAutomated}. This background image was subtracted from the initial image to produce a preprocessed image (\autoref{figure-binarization-concept}, img\_diff).
Next, the core and boundary regions of the flocs were binarized using different methods. For the boundary regions, eroded and dilated images were derived from the preprocessed image, and their difference was binarized using a fixed threshold (\autoref{figure-binarization-concept}, img\_boundaries). For the core regions, adaptive histogram equalization (CLAHE) was applied to the preprocessed image to obtain an enhanced image (\autoref{figure-binarization-concept}, img\_diff\_enhance), which was then binarized using a fixed threshold (\autoref{figure-binarization-concept}, img\_core).
Finally, the binarized images were combined and post-processed to produce the final binarized image (\autoref{figure-binarization-concept}, img\_bin\_result).

After binarization, morphological parameters were calculated for each floc.
The parameters included area ($A$), perimeter ($P$), form factor (Eq. (\ref{definition-form-factor})), convexity (Eq. (\ref{definition-convexity})), solidity (Eq. (\ref{definition-solidity})), aspect ratio (Eq. (\ref{definition-aspect-ratio})), compactness (Eq. (\ref{definition-compactness})), roundness (Eq. (\ref{definition-roundness})), eccentricity \cite{Khan-2018-GeneralizedClassificationModelingActivated}, and extent \cite{Khan-2018-GeneralizedClassificationModelingActivated}.
In Eq. (\ref{definition-convexity}), $A_\text{Conv}$ denotes the area of the convex hull, while in Eq. (\ref{definition-solidity}), $P_\text{Conv}$ denotes its perimeter.
These parameters were defined according to previous studies \cite{Grijspeerdt-1997-ImageAnalysisEstimateSettleability,Amaral-2005-ActivatedSludgeMonitoringWastewater,Mesquita-2009-CorrelationSludgeSettlingAbility,Khan-2018-GeneralizedClassificationModelingActivated} and were computed using the \texttt{regionprops} function in MATLAB.

\subsubsection{Floc classification based on morphological parameters using machine learning}
Flocs extracted from microscopic images were classified using LightGBM \cite{Ke-2017-LightGBMHighlyEfficientGradient} based on morphological parameters.
We used approximately 200 images per sample for training and 100 images per sample for evaluation.
For each classification instance, $n$ flocs ($10 \leqq n \leqq 100$), corresponding to $10n$ morphological parameter values, were randomly selected without replacement to account for variation in the number of flocs among images, and the selected flocs were used to classify the samples.
For each value of $n$, 2000 classification instances per sample were generated for training, and 50 classification instances per sample were generated for evaluation.
20 \% of the training data were used for validation, and training was stopped when the validation performance did not improve for 100 iterations.
Each value of $n$ was evaluated in three independent runs.

\subsection{Frequency analysis of images}
We conducted frequency analysis of the microscopic images to further investigate their image characteristics. Two microscopic images per sample were randomly selected from the deep learning training dataset, and their power spectra were calculated. Details of the implementation are provided in Text S1.
In the implementation,
Python 3.13.12 was used as the programming language.
The following packages were employed: PyTorch 2.13.0 (CPU), torchvision 0.28.0, numpy 2.5.1, matplotlib 3.11.1, and opencv-python 5.0.0.93.

\section{Results and discussion}
\subsection{Model architecture, model size, training strategy, image size}
\begin{figure}[hbtp]
\centering
\includegraphics[width = 12cm]{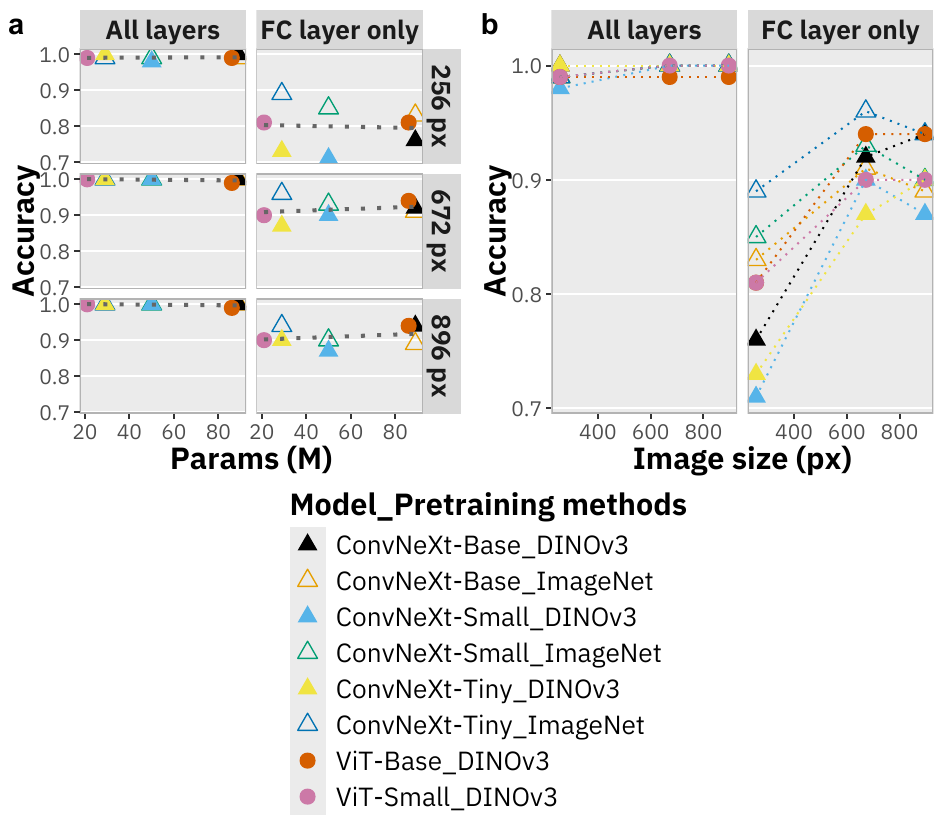}
\caption{Accuracy of the 48 models evaluated in this study: (a) Separate panels are shown for each input image size and training strategy, with either all layers or only the FC layer unfrozen. The horizontal axis in each panel represents model size. The dashed lines indicate linear regression fits; (b) Separate panels are shown for the two training strategies, with either all layers or only the FC layer unfrozen. The horizontal axis in each panel represents the input image size.}
\label{graph_model_image}
\end{figure}

The performance of our 48 models are shown in \autoref{graph_model_image}.
In \autoref{graph_model_image}a, the accuracy results are shown separately for each input image size and training strategy, with either all layers or only the FC layer unfrozen.
Each panel shows results obtained using the ConvNeXt and ViT architectures, with multiple model sizes evaluated for each architecture.
In all panels, the regression lines are nearly horizontal, with $r$ ranging from -0.43 to 0.098 for all layers unfrozen and from -0.056 to 0.248 for the FC layer only unfrozen. These results indicate that the achieved accuracy was largely independent of both model architecture and model size (\autoref{graph_model_image}a).
Meanwhile, the results show that the training strategy influenced model performance, with accuracy ranging from 0.98 to 1.00 when all layers were unfrozen and from 0.71 to 0.96 when only the FC layers were unfrozen (\autoref{graph_model_image}a).
These results suggest that ViT are effective for microscopic image analysis of activated sludge, that increasing model size does not necessarily improve accuracy, and that unfreezing all layers yields better performance than unfreezing only the FC layer.

Our findings regarding model size and training strategy are consistent with previous studies \cite{Borzooei-2024-EvaluationActivatedSludgeSettling,Gao-2025-AIDrivenEarlyWarningSludge}.
A previous study \cite{Borzooei-2024-EvaluationActivatedSludgeSettling} that predicted sludge settleability from microscopic images reported better performance for ConvNeXt-Nano than for ConvNeXt-S.
The lack of improvement in performance with increasing model size may be attributed to the limited amount of data available to adequately train larger models.
A previous study \cite{Gao-2025-AIDrivenEarlyWarningSludge} that detected sludge bulking from stained microscopic images reported higher performance when all layers were unfrozen than when only the FC layers were unfrozen.
When only the FC layers are unfrozen, the feature extraction process remains fixed during training, whereas unfreezing all layers may allow the feature representations themselves to be further optimized. Because the models used in this study were pretrained on images distinct from microscopic images of activated sludge, such as those from ImageNet \cite{Deng-2009-ImageNetLargescaleHierarchicalImage} or Instagram \cite{Siméoni-2025-DINOv3}, training with all layers unfrozen may have enabled the models to learn feature representations more suitable for activated sludge images.
In contrast, no previous studies have reported the effectiveness of ViT for microscopic image analysis of activated sludge, although more than five years have passed since its introduction \cite{Dosovitskiy-2021-ImageWorth16x16Words} and several subsequent studies have applied deep learning to this field \cite{Borzooei-2024-EvaluationActivatedSludgeSettling,Bähr-2025-IntroducingSituActivatedSludge,Li-2025-RealtimeQuantificationActivatedSludge,Gao-2025-AIDrivenEarlyWarningSludge}.
There are two possible reasons why ViT achieved performance comparable to that of ConvNeXt in this study.
First, we used pretrained models. In general, ViTs require more training data than CNNs \cite{Dosovitskiy-2021-ImageWorth16x16Words}, and high accuracy might not have been achieved if the models had been trained from scratch using only microscopic images of activated sludge.
Second, as described in the Materials and Methods, we used ViTs equipped with register tokens \cite{Darcet-2023-VisionTransformersNeedRegisters}, which were not included in the original ViT architecture \cite{Dosovitskiy-2021-ImageWorth16x16Words}. According to the developers of register tokens \cite{Darcet-2023-VisionTransformersNeedRegisters}, their introduction produces smoother feature maps by reducing artifacts. This effect may have facilitated effective transfer learning in the present study.

\autoref{graph_model_image}b shows the effect of image size.
When all layers were unfrozen, accuracy remained high across all image sizes, with no clear effect of image size. In contrast, when only the FC layers were unfrozen, performance varied with image size. Specifically, accuracy differed significantly between image sizes of 256 and 672 px (paired t-test, $p = 0.00013$, Cohen's $d = 1.8$), whereas no significant difference was observed between 672 and 896 px (paired t-test, $p = 0.46$, Cohen's $d = -0.23$).

Because larger images contain more information, deep learning performance would be expected to improve with increasing image size. Consistent with this expectation, accuracy increased when the image size was increased from 256 to 672 px. This results suggests that a sufficiently large image size is beneficial. Interestingly, however, increasing the image size further from 672 to 896 px did not result in a significant improvement in accuracy. Moreover, based on the effect size, performance tended to decrease rather than improve.
These results suggest that, in deep learning-based microscopic image analysis of activated sludge, increasing image size improves accuracy up to a certain point, beyond which further increases provide little or no additional benefit.
A previous study \cite{Wang-2024-PredictionActivatedSludgeSedimentation} used deep learning to classify microscopic images based on sludge settleability and compared several image sizes (224, 299, 448, and 512 px). The study showed that accuracy improved as image size increased, but that 448 px was sufficient to achieve high accuracy. Their findings align with our results.

\subsection{Pretraining methods}

\begin{figure}[hbtp]
\centering
\includegraphics[width = 12cm]{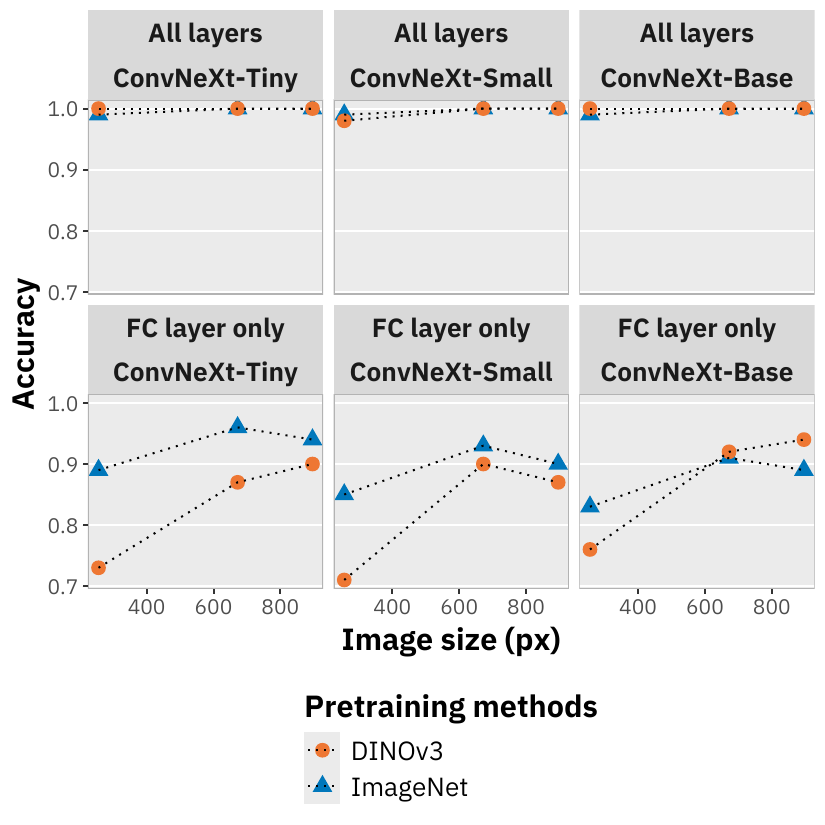}
\caption{Differences in accuracy achieved by our ConvNeXt models across pretraining methods.}
\label{graph_pretraining_methods}
\end{figure}

\autoref{graph_pretraining_methods} shows the difference in accuracy acieved across pretraining methods.
Although two pretraining methods were evaluated in this study, no significant difference in accuracy was observed between them (paired t-test, $p = 0.050$, Cohen's $d = 0.24$) .
This result suggests that pretraining with DINOv3 is effective for microscopic image analysis of activated sludge, as is supervised pretraining on ImageNet, which has been widely used in previous studies \cite{Satoh-2021-DeepLearningbasedMorphologyClassification,Borzooei-2024-EvaluationActivatedSludgeSettling,Li-2025-RealtimeQuantificationActivatedSludge}.

The increase in accuracy with increasing image size appeared to be somewhat more pronounced with DINOv3 pretraining than with ImageNet pretraining (\autoref{graph_pretraining_methods}). The original DINOv3 study \cite{Siméoni-2025-DINOv3} also reported improved performance with increasing image size, despite differences in the tasks evaluated.

\subsection{Field of view vs resolution}

\begin{figure}[hbtp]
\centering
\includegraphics[width = 12cm]{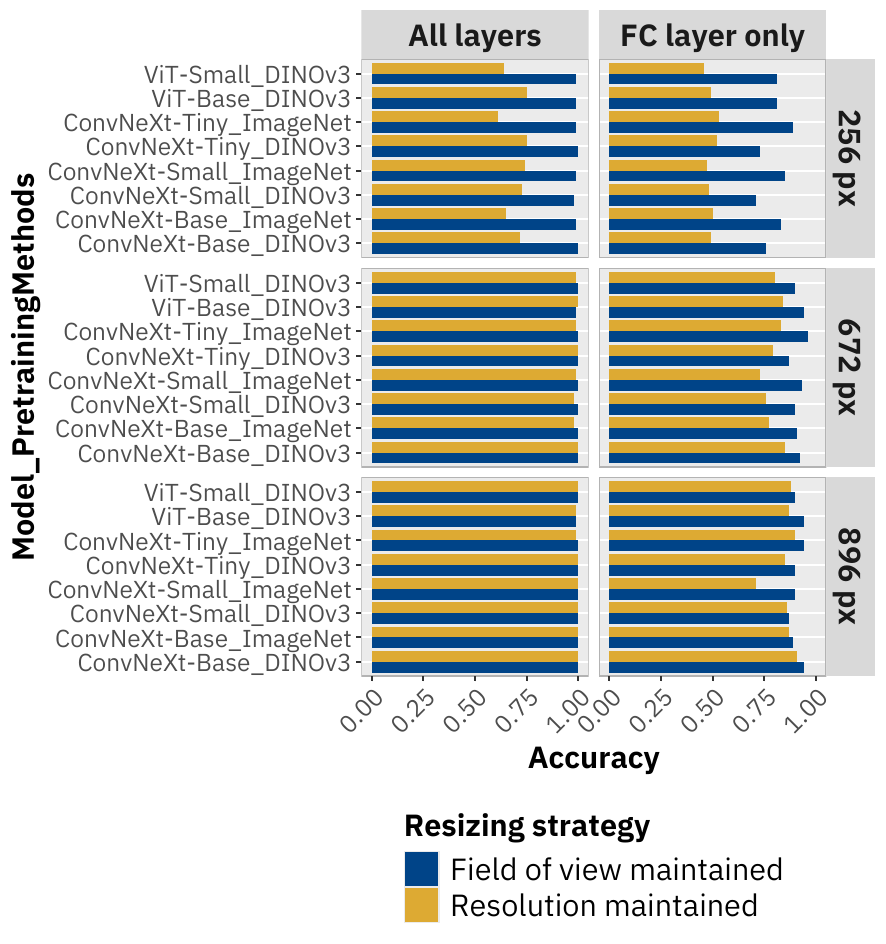}
\caption{Comparison of accuracy between resizing strategies that maintain the field of view and resolution. Accuracy obtained from our 96 trained models is shown in separate panels for each image size and training strategy. Bar length represents accuracy. Blue bars indicate results obtained when the field of view was maintained, whereas yellow bars indicate results obtained when the resolution was maintained.}
\label{graph_view_resolution}
\end{figure}

\autoref{graph_view_resolution} compares the performance of our models when either the field of view or the resolution was maintained.
We trained 48 models under each condition, for a total of 96 models. Maintaining the field of view resulted in higher accuracy than maintaining the resolution for most of the 48 model settings, except for several settings in which accuracy was already close to 1 (paired t-test, $p = 2.1\times10^{-8}$, Cohen's $d = 0.70$).
These results suggest that, when preprocessing activated sludge microscopic images for deep learning-based analysis, maintaining the field of view is more important than maintaiing high image resolution.

Several possible explanations can be considered for why maintaining the field of view was more effective than maintaining the resolution in improving accuracy.
One possible explanation is that morphology played a more important role than texture in the benchmark classification.
Maintaining resolution is likely advantageous for evaluating floc surface texture.
However, reducing the field of view to maintain resolution makes it more difficult to capture the overall floc morphology.
If this hypothesis is correct, applying activated sludge microscopic images to tasks where texture is more important than morphology (e.g., detection of protozoa and metazoa) could yield results opposite to those observed in this study.
A second possible explanation is a mismatch between the field of view and floc size.
Under the conditions of this study, maintaining the field of view yielded a side length of 747 \textmu m, whereas reducing the field of view to maintain resolution at an image size of 256 px yielded a side length of only 158 \textmu m.
Although floc size distributions likely vary among the samples, Wil\'{e}n and Balmer \cite{Wilén-1999-EffectDissolvedOxygenConcentration} reported that floc sizes followed a log-normal distribution ranging from 11.6 to 1128 \textmu m.
In addition, QIA in the present study detected flocs with an equivalent diameter of up to 721 \textmu m (Fig. S2).
Taken together, these observations imply that a field of view of 158 \textmu m may have been too small.
Another possible explanation is the reduction in information content. When the resolution is maintained while the field of view is reduced, both the representativeness and information content of the image decrease. In contrast, when the field of view is maintained while the resolution is reduced, the amount of information in the image may not decrease substantially. To examine this possibility, we conducted a frequency analysis, as shown in Fig. S3. We calculated the power spectra of our microscopic images and found that the image information was concentrated primarily in the low-frequency components. These results indicate that low-frequency components dominate microscopic images of activated sludge and that relatively little information is lost when the resolution is reduced. We assumed that recent improvements in image sensor quality and pixel count have exceeded the resolving capability of microscope objective lenses. Consequently, useful information in the images may be concentrated primarily in lower spatial frequencies, and the effect of reducing image resolution may therefore be limited, as in the case of empty magnification.

\subsection{Comparison of deep learning and quantitative image analysis}
\begin{figure}[hbt]
\centering
\includegraphics[width = 11cm]{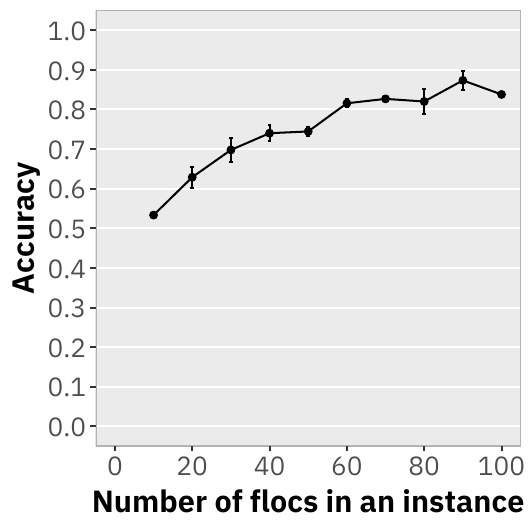}
\caption{Relationship between the number of flocs per instance and classification accuracy in benchmark tasks using QIA-derived morphological parameters and machine learning. The error bars indicate the standard error.}
\label{graph_result_3}
\end{figure}

We binarized the microscopic images acquired for QIA, extracted an average of 3.79 flocs per image, and calculated 10 morphological parameters for each extracted floc.
For benchmark classification, the morphological parameters derived from $n$ flocs were treated as a single instance.
\autoref{graph_result_3} shows the relationship between classification accuracy and the value of $n$.
When more than 60 flocs were included in each instance, the classification accuracy exceeded 0.8, reaching a maximum of 0.873 at $n = 90$ (\autoref{graph_result_3}).
However, from two perspectives, our results indicate that deep learning outperformed the QIA-based benchmark classification. First, deep learning achieved an accuracy above 0.9 (\autoref{graph_model_image}) and showed superior classification performance compared with the QIA-based approach. Second, deep learning was substantially more efficient in terms of the amount of image data required. Whereas deep learning treated a single image as one instance, the QIA-based approach required more than 60 flocs per instance to achieve an accuracy above 0.8. This requirement of more than 60 flocs corresponded to approximately 16 images.

\subsection{Limitation and suggestion}
\subsubsection{Limitation of our research}
This study has two limitations.
The first limitation is the lack of discussion regarding interpretability.
Throughout this study, we consistently used accuracy as the metric for evaluating the performance of deep learning and QIA. However, interpretability, which was not discussed in this study, is also an important consideration, particularly when comparing deep learning with QIA.
In general, deep learning achieves outstanding performance, but it is often difficult to interpret how deep learning models process and evaluate images. Recently, several visualization methods have been developed \cite{Selvaraju-2017-GradCAMVisualExplanationsDeep,Chattopadhyay-2018-GradCAMImprovedVisualExplanations,Muhammad-2020-EigenCAMClassActivationMap} and applied to microscopic image analysis of activated sludge \cite{Borzooei-2024-EvaluationActivatedSludgeSettling}. However, the resulting interpretations have generally been qualitative and subjective rather than quantitative and objective. In contrast, morphological parameters directly represent floc morphology and are therefore relatively easy to interpret. Indeed, in our previous study \cite{Hakoshima-2026-NewInsightsFlocMorphology}, we used QIA because our objective was to investigate the relationship between floc morphology and the microbial community. In the present study, our results indicate that deep learning outperformed the QIA-based benchmark classification, but this conclusion was based on accuracy and data requirements rather than interpretability.

The second limitation is the limited generalizability of our findings. In this study, we used a benchmark classification task in which three types of activated sludge samples were classified based on microscopic images. However, this task does not fully reflect practical applications, where the objective is often to predict sludge properties or process performance. Therefore, further evaluation is needed to determine whether the findings of this study can be generalized to such prediction tasks.
We hypothesize that tasks in which global image features play a dominant role may show trends similar to those observed in this study, whereas tasks that depend primarily on local image features may yield different results. The classification task examined here was likely driven mainly by global features of activated sludge images. Similar trends may therefore be observed in tasks such as predicting sludge settleability, for which global image characteristics are also likely to be important.
Indeed, the trends observed in our study for model size, image size, and training strategies were consistent with those reported in previous studies \cite{Borzooei-2024-EvaluationActivatedSludgeSettling,Gao-2025-AIDrivenEarlyWarningSludge,Wang-2024-PredictionActivatedSludgeSedimentation} (\autoref{graph_model_image}).
In contrast, tasks such as identifying bacterial species based on floc texture may rely more strongly on local features, and the trends may therefore differ from those observed in this study.

\subsubsection{Suggestion of our research}
This study's findings provide several useful insights for future research on microscopic image analysis of activated sludge. First, our results showed that increasing model size does not necessarily improve deep learning performance and that unfreezing all layers yields better performance than unfreezing only the FC layer. These trends have also been indicated in a previous study aimed at detecting sludge bulking \cite{Borzooei-2024-EvaluationActivatedSludgeSettling,Gao-2025-AIDrivenEarlyWarningSludge}. The similar trends observed in the present study, which addressed classification among different activated sludge samples, suggest that these findings may be broadly applicable to deep learning-based microscopic image analysis of activated sludge.

Second, we found that models pretrained using methods other than supervised learning on ImageNet achieved accuracy comparable to that of models pretrained using conventional supervised learning on ImageNet. We also found that ViTs performed comparably to CNNs. Previous studies on microscopic image analysis of activated sludge \cite{Satoh-2021-DeepLearningbasedMorphologyClassification,Borzooei-2024-EvaluationActivatedSludgeSettling,Li-2025-RealtimeQuantificationActivatedSludge,Wang-2024-PredictionActivatedSludgeSedimentation} have generally relied on CNNs pretrained through supervised learning on ImageNet. In the broader field of deep learning-based image recognition, however, research has increasingly focused on developing general-purpose foundation models \cite{Li-2023-MultimodalFoundationModelsSpecialists} using transformer architectures and self-supervised learning on large-scale datasets beyond ImageNet \cite{Chen-2020-SimpleFrameworkContrastiveLearning,He-2020-MomentumContrastUnsupervisedVisual,Siméoni-2025-DINOv3}. In light of these developments, we propose that environmental engineering research should also explore ViTs and alternative pretraining approaches to keep pace with advances in AI.

Third, across image size, resolution, and field of view, our results show that maintaining the field of view and using a sufficiently large image size are important for deep learning-based microscopic image analysis of activated sludge.
In contrast, maintaining the original resolution or using the maximum possible image size does not appear to be necessary. Reducing image size helps increase computational efficiency and supports model training when computational resources are limited. Therefore, these findings may provide practical guidance for future studies.
An important difference between the present study and previous study \cite{Wang-2024-PredictionActivatedSludgeSedimentation} that reported similar trends is that we further investigated the possible reasons for these observations. Our frequency analysis showed that microscopic images of activated sludge contain predominantly low-spatial-frequency information. We suggest that this may be because the number of pixels provided by modern image sensors exceeds the resolving capability of the microscope objective lens. Although objective-lens resolving power is constrained by physical principles, image-sensor pixel counts are likely to continue increasing. Based on our findings, we therefore propose that further increases in image sensor pixel count alone are unlikely to substantially improve the performance of deep learning-based microscopic image analysis of activated sludge.

Finally, our study quantitatively showed that deep learning outperforms QIA in predictive performance. Previous studies \cite{Satoh-2021-DeepLearningbasedMorphologyClassification,Borzooei-2024-EvaluationActivatedSludgeSettling,Gao-2025-AIDrivenEarlyWarningSludge,Wang-2024-PredictionActivatedSludgeSedimentation} have mainly justified using deep learning over QIA on qualitative grounds, such as the labor-intensive and time-consuming image preprocessing required for QIA. We agree that these are important advantages of deep learning. However, we believe a method's ability to serve as a powerful image analysis tool matters more than whether it reduces labor or processing time. As discussed in the limitations, the present study did not evaluate model interpretability. Nevertheless, when high predictive performance is the primary objective, our results suggest preferring deep learning over QIA for microscopic image analysis of activated sludge.

\section{Conclusion}
This research examines how different training conditions affect deep learning performance in microscopic image analysis of activated sludge and compares deep learning with QIA. We prepared three types of activated sludge samples, classified their microscopic images, and evaluated classification accuracy. Our results showed that increasing model size did not necessarily improve performance and that transformer-based architectures and alternative pretraining methods were also effective for microscopic image analysis of activated sludge. We further examined how image size, field of view, and resolution affected accuracy. The results indicated that using images that were too small reduced performance, whereas increasing image size beyond a certain level did not lead to further improvement. In addition, maintaiing the field of view is more improtant than maintaining the resolution. We also found that the information in microscopic images of activated sludge was concentrated primarily in low-frequency components, which may explain these observations. This finding further suggests that continued increases in image sensor pixel counts alone may not substantially improve the performance of microscopic image analysis of activated sludge. Finally, we quantitatively showed that deep learning outperformed QIA in terms of accuracy. These results provide an additional rationale for applying deep learning to microscopic image analysis of activated sludge. Our findings offer practical guidance for future studies.

\section*{Data availability}
Resarch data are available from the corresponding author upon request, with approval from the municipalities where the samples were collected.

\section*{Acknowledgements}
Part of this work was supported by JSPS KAKENHI Grant Number
JP24KJ0636, JP25K22111, and JP26K24641.]
Part of this work was supported by a collaborative research project by the University of Tokyo and KUBOTA Corporation.
The authors thank Ms. Tomoko INOUE (technical staff) for assistance in experiments.
The authors also acknowledge the Division of Creative Activity, Global Center in Engineering Education Institute for Innovation in International Engineering Education, Graduate School of Engineering, The University of Tokyo, for their support in the developments of research equipment.

\appendix
\section{Definition of parameters}
\begin{align}\label{definition-form-factor}
\text{Form factor} &= \frac{4\pi A}{P^2}\\\label{definition-convexity}
\text{Convexity} &= \frac{A}{A_\text{Conv}}\\\label{definition-solidity}
\text{Solidity} &= \frac{P_\text{Conv}}{P} \\\label{definition-aspect-ratio}
\text{Aspect ratio} &= \frac{\text{Minor axis length}}{\text{Major axis length}}\\\label{definition-compactness}
\text{Compactness} &= \frac{\sqrt{\frac{4A}{\pi}}}{\text{Major axis length}} \\\label{definition-roundness}
\text{Roundness} &= \frac{4A}{\pi\text{Major axis length}^2}
\end{align}

\section{Supplementary figures}
\begin{figure}[h]
\centering
\includegraphics[width = 9cm]{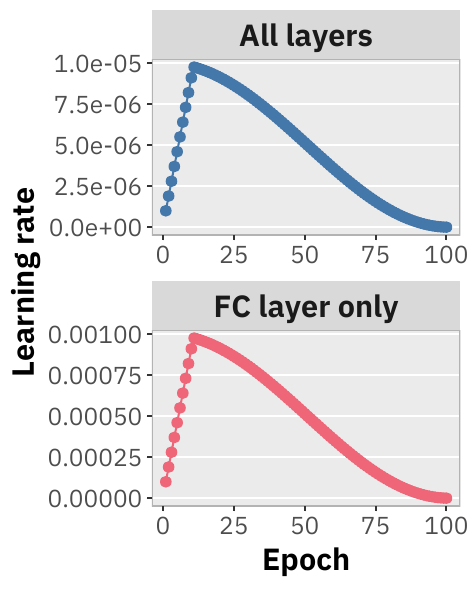}
\caption{Learning rate schedule over training epochs. We used different learning rates depending on whether all layers or only the FC layer were unfrozen.}
\end{figure}


\begin{figure}[h]
\centering
\includegraphics[width = 11cm]{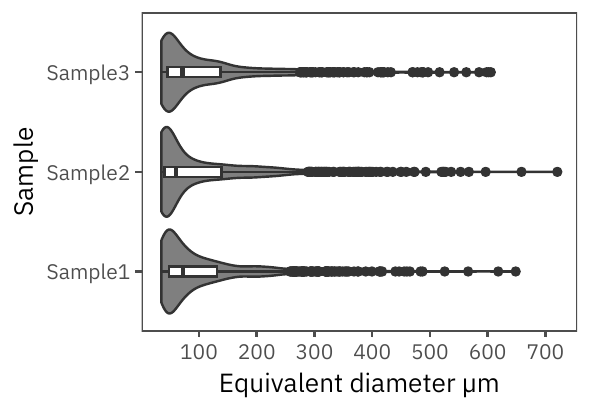}
\caption{Floc size distribution.}
\end{figure}


\begin{figure}[h]
\centering
\includegraphics[width = 8cm]{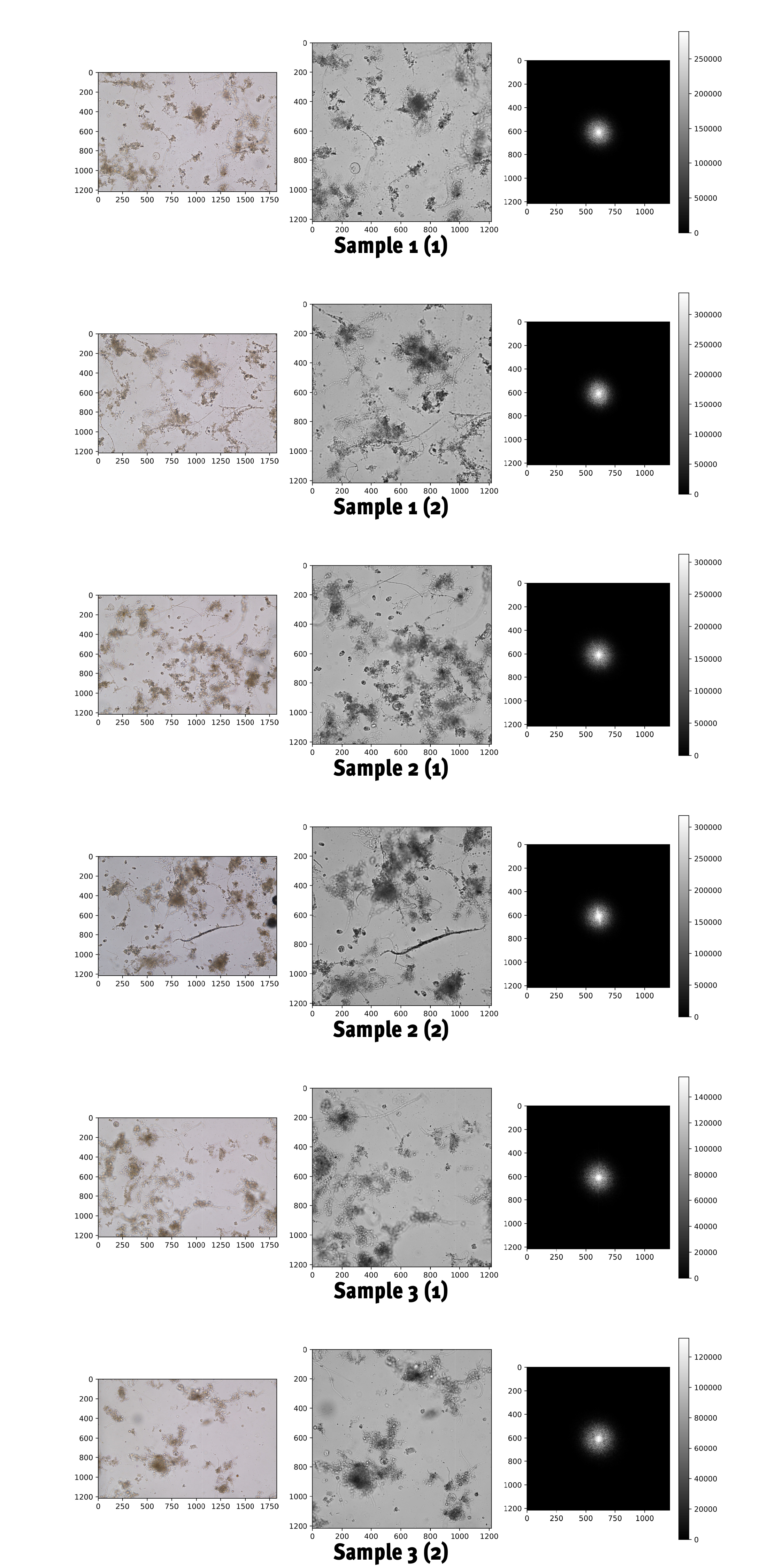}
\caption{Results of frequency analysis. The left, middle, and right panels show the original image, the image after center cropping and conversion to grayscale, and the power spectrum, respectively. The power spectrum is shifted so that the low-frequency components are located at the center.}
\end{figure}

\clearpage
\section{Supplementary texts}
\lstset{
    language = Python,
    breaklines = true,
    breakindent = 12pt,
    numbers = left,
    captionpos = t,
    frame = tb
}
\begin{lstlisting}[caption=The implementation of frequency analysis. The paths of the folder was erased.]
import os
import random
from numpy import zeros_like
import torch
from torchvision import transforms as transforms
import numpy as np
import matplotlib.pyplot as plt
import cv2

random.seed(123)


class FourierTransformGray_Power:
    def __call__(self, img):
        f = self.FourierTransform(img)
        PowerSpectrum = torch.abs(f) ** 2
        return PowerSpectrum

    def FourierTransform(self, img):
        f = torch.fft.fft2(img)
        fshift = torch.fft.fftshift(f)
        return fshift


def RGB2FLOAT(img):
    return np.float64(img) / 255


Power = FourierTransformGray_Power()
transform_preProcessing = transforms.Compose(
    [
        transforms.ToTensor(),
        transforms.CenterCrop(1216),
    ]
)
transform_power = transforms.Compose(
    [
        transform_preProcessing,
        Power,
    ]
)

def image_read(path):
    img = cv2.imread(path)
    img = cv2.cvtColor(img, cv2.COLOR_BGR2RGB)
    img_rgb = RGB2FLOAT(img)
    img_gray = cv2.cvtColor(img, cv2.COLOR_RGB2GRAY)
    return img_rgb, img_gray


def image_processing(img_rgb,img_gray):
    img_gray_resize = transform_preProcessing(img_gray)
    img_gray_resize = img_gray_resize.to("cpu").detach().numpy().copy()
    img_gray_resize = np.squeeze(img_gray_resize, axis=0)
    img_gray_power = transform_power(img_gray)
    img_gray_power = img_gray_power.to("cpu").detach().numpy().copy()
    img_gray_power = np.squeeze(img_gray_power, axis=0)

    fig, axes = plt.subplots(1, 3, figsize=(15, 5))
    axes[0].imshow(img_rgb)
    axes[1].imshow(img_gray_resize, cmap="gray")
    im = axes[2].imshow(
        img_gray_power,
        cmap="gray",
        vmin=0,
        vmax=np.percentile(img_gray_power, 99.5),
    )
    fig.colorbar(im, ax=axes[2])


folder_path_S1 = 
files_S1 = os.listdir(folder_path_S1)
img_S1_1_name = random.choice(files_S1)
img_S1_1_name = folder_path_S1 + "/" + img_S1_1_name
img_S1_1_rgb, img_S1_1_gray = image_read(img_S1_1_name)
img_S1_2_name = random.choice(files_S1)
img_S1_2_name = folder_path_S1 + "/" + img_S1_2_name
img_S1_2_rgb, img_S1_2_gray = image_read(img_S1_2_name)


folder_path_S2 = 
files_S2 = os.listdir(folder_path_S2)
img_S2_1_name = random.choice(files_S2)
img_S2_1_name = folder_path_S2 + "/" + img_S2_1_name
img_S2_1_rgb, img_S2_1_gray = image_read(img_S2_1_name)
img_S2_2_name = random.choice(files_S2)
img_S2_2_name = folder_path_S2 + "/" + img_S2_2_name
img_S2_2_rgb, img_S2_2_gray = image_read(img_S2_2_name)

folder_path_S3 =
files_S3 = os.listdir(folder_path_S3)
img_S3_1_name = random.choice(files_S3)
img_S3_1_name = folder_path_S3 + "/" + img_S3_1_name
img_S3_1_rgb, img_S3_1_gray = image_read(img_S3_1_name)
img_S3_2_name = random.choice(files_S3)
img_S3_2_name = folder_path_S3 + "/" + img_S3_2_name
img_S3_2_rgb, img_S3_2_gray = image_read(img_S3_2_name)

image_processing(img_S1_1_rgb, img_S1_1_gray)
plt.savefig("img_S1_1.png",dpi = 900)
plt.clf()
image_processing(img_S1_2_rgb, img_S1_2_gray)
plt.savefig("img_S1_2.png", dpi=900)
plt.clf()

image_processing(img_S2_1_rgb, img_S2_1_gray)
plt.savefig("img_S2_1.png", dpi=900)
plt.clf()
image_processing(img_S2_2_rgb, img_S2_2_gray)
plt.savefig("img_S2_2.png", dpi=900)
plt.clf()

image_processing(img_S3_1_rgb, img_S3_1_gray)
plt.savefig("img_S3_1.png", dpi=900)
plt.clf()
image_processing(img_S3_2_rgb, img_S3_2_gray)
plt.savefig("img_S3_2.png", dpi=900)
plt.clf()
\end{lstlisting}

\printbibliography

\end{document}